\documentclass[sigconf]{acmart}
\usepackage{algorithm}%
\usepackage{algorithmicx}%
\usepackage{algpseudocode}%
\usepackage{makecell}%
\usepackage{multirow}%
\usepackage{float}
\AtBeginDocument{%
  }

\copyrightyear{2024} 
\acmYear{2024} 
\setcopyright{acmlicensed}\acmConference[GECCO '24 Companion]{Genetic and Evolutionary Computation Conference}{July 14--18, 2024}{Melbourne, VIC, Australia}
\acmBooktitle{Genetic and Evolutionary Computation Conference (GECCO '24 Companion), July 14--18, 2024, Melbourne, VIC, Australia}
\acmDOI{10.1145/3638530.3664128}
\acmISBN{979-8-4007-0495-6/24/07}

\begin{document}

\title{Deep Symbolic Optimization for Combinatorial Optimization: \\Accelerating Node Selection by  Discovering Potential Heuristics}

\author{Hongyu Liu}
\email{horin@mail.ustc.edu.cn}
\orcid{0009-0008-8036-9276}
\affiliation{%
  \institution{University of Science and Technology of China}
  \city{Hefei}
  \country{China}
}

\author{Haoyang Liu}
\email{dgyoung@mail.ustc.edu.cn}
\orcid{0009-0002-8594-4045}
\affiliation{%
  \institution{University of Science and Technology of China}
  \city{Hefei}
  \country{China}
}

\author{Yufei Kuang}
\email{yfkuang@mail.ustc.edu.cn}
\orcid{0009-0000-2483-3638}
\affiliation{%
  \institution{University of Science and Technology of China}
  \city{Hefei}
  \country{China}
}

\author{Jie Wang}
\email{jiewangx@ustc.edu.cn}
\orcid{0000-0001-9902-5723}
\affiliation{%
  \institution{University of Science and Technology of China}
  \city{Hefei}
  \country{China}
}

\author{Bin Li}
\authornote{Corresponding author}
\email{binli@ustc.edu.cn}
\orcid{0000-0002-2332-3959}
\affiliation{%
  \institution{University of Science and Technology of China}
  \city{Hefei}
  \country{China}
}


\begin{abstract}
    Combinatorial optimization (CO) is one of the most fundamental mathematical models in real-world applications.
  Traditional CO solvers, such as Branch-and-Bound (B\&B) solvers, heavily rely on expert-designed heuristics, which are reliable but require substantial manual tuning. 
  Recent studies have leveraged deep learning (DL) models as an alternative to capture rich feature patterns for improved performance on GPU machines.
  Nonetheless, the drawbacks of high training and inference costs, as well as limited interpretability, severely hinder the adoption of DL methods in real-world applications.
  To address these challenges, we propose a novel deep symbolic optimization learning framework that combines their advantages.
  Specifically, we focus on the node selection module within B\&B solvers---namely, deep symbolic optimization for node selection (Dso4NS). 
  With data-driven approaches, Dso4NS guides the search for mathematical expressions within the high-dimensional discrete symbolic space and then incorporates the highest-performing mathematical expressions into a solver.
  The data-driven model captures the rich feature information in the input data and generates symbolic expressions, while the expressions deployed in solvers enable fast inference with high interpretability.
  Experiments demonstrate the effectiveness of Dso4NS in learning high-quality expressions, outperforming existing approaches on a CPU machine.
  Encouragingly, the learned CPU-based policies consistently achieve performance comparable to state-of-the-art GPU-based approaches.
\end{abstract}




\maketitle

\section{Introduction}
Combinatorial optimization (CO) holds a fundamental position within the field of mathematical optimization (MO) due to its wide range of applications in real-world scenarios, including route planning \cite{liu2008tsp}, vehicle scheduling \cite{chen2010integrated}, chip design \cite{ma2019accelerating} and so on. 
Due to the NP-hard nature, there is generally no known polynomial-time algorithm to solve CO optimally. 
During the solving process, the solvers are required to deal with discrete decision variables with exponentially large and highly complex search spaces, leading to expensive computational and time costs.

Existing traditional CO solvers, such as SCIP \cite{achterberg2007constraint} and Gurobi \cite{optimization2021llc}, rely heavily on expert-designed heuristics or rule-based working flows. 
Among these heuristics, heuristics in the form of mathematical expressions are easy to deploy in the solver with fast inference speed and low computational cost.
As is designed by human experts, the mathematical expressions are coincident with human intuitions
and thus regarded to be reliable.
However, designing these heuristics needs considerable engineering experience and extensive manual tuning, which is time-consuming and challenging in many real-world applications.
Moreover, these heuristics do not consider specific data distribution of the input features thus limiting  further improvement for higher performance. 

As a consequence, recent works attempt to leverage machine learning (ML) to improve the solving efficiency of CO solvers. 
Learning strategies from past information and data-driven patterns, the ML approaches utilize the deep model to take the place of these heuristics on decision-making behaviors such as primal heuristics \cite{paulus2024learning}, branching \cite{gasse2019exact}, and node selection \cite{he2014learning}.
Though promising improvement, the emergence of training and deployment issues impedes the wide applications of the learning-based solvers. 
First, common deep learning methods require a considerable amount of data for training, but in real-world scenarios, the scarcity and privacy of training data strongly restrict development in specific fields. 
Second, as deep models grow in depth with more complex model structures, the demand for computing resources rises rapidly, especially high-end  GPU resources. 
Considering the actual situation of pure CPU-based machines in industrial proposes, deployment and inference of deep models may lead to a significant increase in inference time.
Furthermore, deep models are often used as black-box tools for decision-making in solvers, thus the lack of reliability and interpretability remains a great challenge.

Therefore, a question arises whether we can combine the advantages of both heuristics and learning-based methods---employing data-driven approaches to learn heuristics with high-quality mathematical expressions---while simultaneously addressing their drawbacks. 
Inspired by the idea of genetic programming \cite{gp} and the recent development of deep symbolic regression \cite{petersen2019deep}, we propose a novel deep symbolic optimization framework (Dso4NS) for the node selection module in B\&B solvers.
We focus on the node selection module since it is one of the most important modules within a B\&B solver, where the performing efficiency and quality of the selected nodes greatly influence the overall solving efficiency.  
Specifically, we leverage a deep sequential model for symbolic regression.
The deep model captures the rich distributional information in the input features and generates symbolic expressions for node selection heuristics, in which the model acts as a decision policy to search for expressions with high performance in the high-dimension and discrete expression space.
For training, we use behavioral cloning accuracy as the fitness measure to perform policy gradient.
Finally, we deploy the learned symbolic expressions in the solver for fast inference on pure CPU-based machines.

We conduct experiments on three challenging NP-hard combinatorial optimization problems.
The results demonstrate that Dso4NS is able to outperform existing node selection baselines, leading to distinct improvement in solving time, number of the explored nodes and the primal-dual integral.
Encouragingly, extensive experiments show that the learned policies evaluated on a purely CPU-based machine consistently achieve comparable performance  to that of the state-of-the-art GPU-based approaches.

\section{Background}
\subsection{MILP and Branch-and-Bound Algorithm}
In practice, a large family of CO problems can be formulated as Mixed Integer Linear Programming (MILP).
A MILP takes the form as follows:
\begin{align*}
    \mathop{\arg\min}\limits_{x} \{ \mathbf{c}^\top\mathbf{x}|\mathbf{Ax}\leq \mathbf{b}, x\in\mathbb{Z}^p\times \mathbb{R}^{n-p} \}
\end{align*}
where $\mathbf{c}$ represents the cost vector, $\mathbf{A}$ denotes the constraint matrix, $\mathbf{b}$ is the constraint right-hand-side vector, and $p$ denotes the number of integer variables. 

The most popular exact solving framework for MILPs is the Branch-and-Bound (B\&B) algorithm, which iteratively decomposes the MILP problem into subproblems and organizes them as nodes within a search tree.

Specifically, the root node of the search tree represents the original MILP problem.
The algorithm then solves the linear program (LP) relaxation of the MILP to obtain the optimal solution $x^{*}_{i}$. 
Suppose the optimal solution contains fractional values for the integer variables. In that case, B\&B introduces constraints $x_i \leq \lfloor x^{*}_{i} \rfloor$ and $x_i \geq \lceil x^{*}_{i} \rceil$ to partition the problem into two subproblems, thereby generating two new child nodes. 
After that, B\&B selects a new child node (subproblem) to explore according to certain node selection heuristics and perform the above decomposition repeatedly.
This process continues until a globally optimal solution is found or the search space is exhausted.

In the B\&B algorithm, efficient and high-quality branching variable selection policy (branching policy) and node selection policy can lead to a significant improvement in the overall solving efficiency.
In this paper, we focus on designing an effective node selection policy via learning-based techniques to better explore the search tree and guide the algorithm toward optimal solutions. 

\subsection{Node Selection Heuristics}
Traditional MILP solvers perform node selection typically according to human-designed heuristics.
Existing node selection heuristics can broadly be categorized into two classes: direct searching approaches and searching with solving statistics. 
Direct searching, such as Depth-First Search (DFS) \cite{Dakin1965ATA} and Breadth-First Search (BFS), is easy to implement but may suffer from low efficiency because of regardless of current solving information. 
Searching with solving statistics, like Best-First Search \cite{1555942} and Node Comparison (\textit{NODECOMP}), takes the current solving information, such as the dual bound, into consideration and leverages adaptive mechanisms to guide the searching process. 
However, the effectiveness of these heuristics strategies heavily relies on considerable manual tuning and expert experiences.

The best estimate search heuristics (Estimate) \cite{benichou1971experiments, forrest1974practical}, based on the \textit{NODECOMP} heuristic, is one of the most popular node selection strategies implemented in state-of-the-art solvers.
Estimate aims to efficiently explore the search space by prioritizing nodes that are estimated to contribute the most towards improving the objective value.
In this approach, an estimation score, derived from pseudo-cost statistics associated with a node in the search tree, indicates the potential increase in the objective value that would result from selecting that node for exploration. 
Estimate then selects the node with the highest estimation score, for which is the most promising candidate.
By focusing on nodes with higher estimated contributions to improving the objective value, the estimate strategy facilitates more effective pruning of unpromising branches and directs the search towards regions of the search space likely to contain optimal solutions.

\subsection{Symbolic Regression}
Unlike conventional regression methods that typically fit a dataset $\{(\mathbf{x}_i,y_i)\}$ using predefined function class (like linear or polynomial equations), symbolic regression searches for more general functional form $f$ that best describes the data in the space of mathematical expressions. 
Since mathematical expressions are powerful tools to describe physical laws and scientific rules, symbolic regression acts as an important data-driven method in scientific discovery \cite{kim2020integration}, decision-making, and control fields \cite{diveev2021machine}.

Regarding the combination of the discrete part (the variable symbols) and continuous part (specific continuous parameters) in the expression space, symbolic regression usually needs to tackle a high-dimension and complex search space. 
Conventional learning methods employ evolutionary algorithms, such as genetic programming (GP) \cite{Koza1992GeneticPO, schmidt2009distill, Baeck2000EvolutionaryC1}, to address the NP-hard search problem. 
In these methods, a fitness measure is carefully designed to evaluate the quality of generated expressions.
For GP-based symbolic regression, a series of evolving operations including recombination and mutation are applied to each generated expression in order to ultimately generate and obtain the one with the highest fitness function. 
However, GP lacks the ability to leverage data distribution characteristics, which significantly limits its performance on specific types of problems.

\section{Method}
\begin{figure*}[h]
  \centering
  \includegraphics[width=\textwidth]{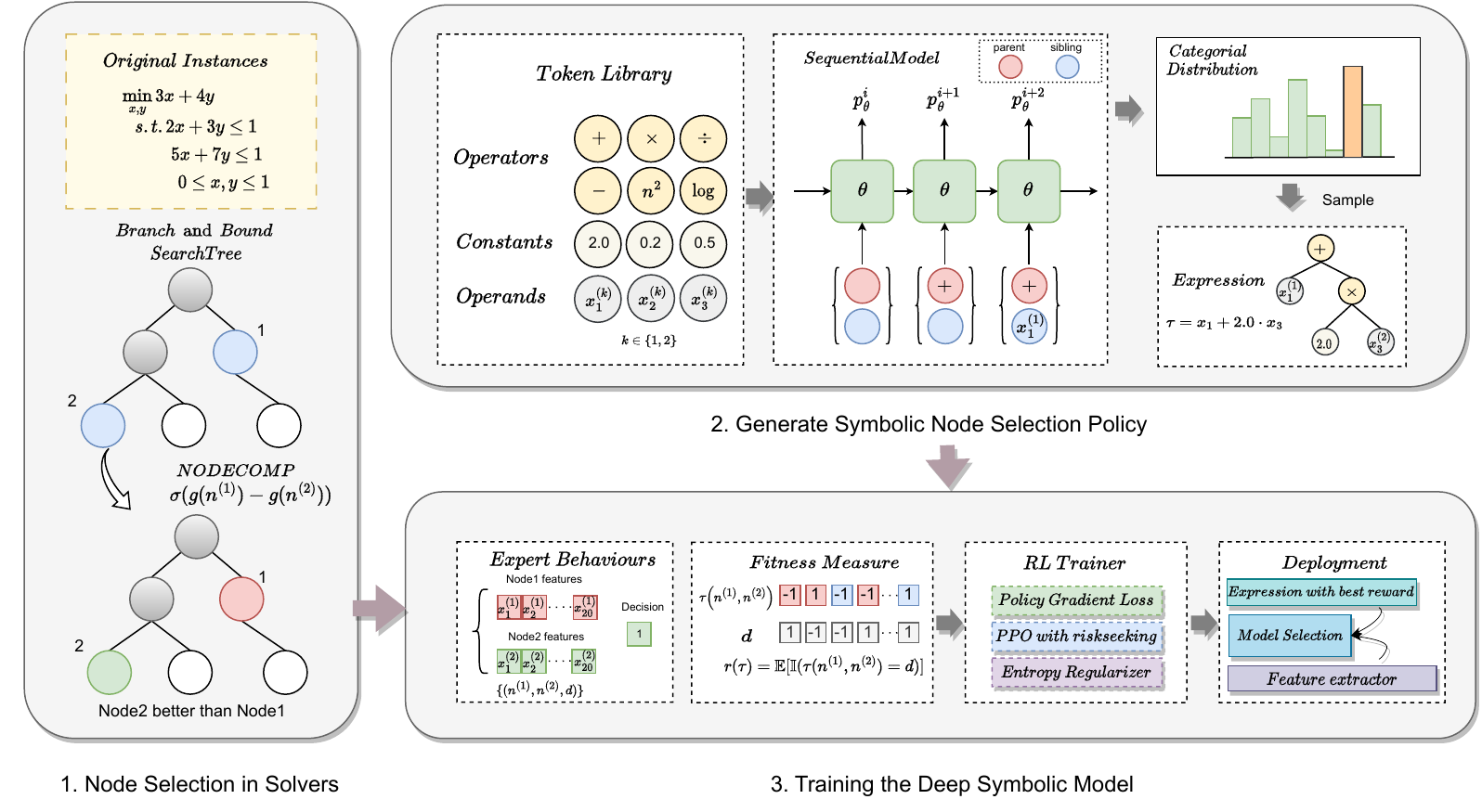}
  \caption{Illustration of the deep symbolic regression framework for heuristic discovery in the node selection module. 
  Part 1 illustrates the \textit{NODECOMP} heuristic for node selection in a B\&B solver. Then, we aim to learn a better mathematical expression for the scoring function $g$.
  Part 2 is the generation process of a symbolic expression, in which we leverage a sequential model to iteratively choose tokens in a predefined token library.
  Part 3 shows the reinforcement-learning-based training algorithm for the expression generator.}
  \label{fig:overview}
\end{figure*}

While traditional heuristics need hand-craft design and deep models lack interpretability, we develop a framework (Dso4NS) that generates mathematical expressions to select nodes in B\&B to tackle these problems.
Dso4NS learns by leveraging \textit{fitness measure} described below. 
Our framework holds two strengths: one is that the symbolic expression form is highly readable and interpretable, and the other is that it can be directly deployed on pure CPU machines for fast inference without the usage of neural networks or GPUs.
The overview of Dso4NS is illustrated in Figure \ref{fig:overview}.

\subsection{Expressions for Node Selection}
The node selection heuristic based on the \textit{NODECOMP} function \cite{labassi2022learning} takes the form of $f(n^{(1)}, n^{(2)})=\sigma(g(n^{(1)})-g(n^{(2)}))$, where $n^{(i)}$ is the node features for $node_i$ $(i=1,2)$, $g:\mathbb{R}^n\rightarrow\mathbb{R}$ is a manually defined scoring function and $\sigma$ is the sigmoid function. 
The \textit{NODECOMP} returns -1 if $f\geq0.5$, i.e. we decide to select $node_1$ to explore next instead of $node_2$, otherwise returns 1 and chooses $node_2$ to explore.  
Our goal is to directly replace function $f$ with the more powerful generated expressions since the input of $f$ contains higher dimensionality than $g$.

Notice that $f$ is symmetric with $f(n^{(1)}, n^{(2)})=1-f(n^{(2)}, n^{(1)})$ while our replacement may lack consistency, we further attempt to replace function $g$ in comparative experiments.

Notice that normal expression sequence, e.g. simply listing the symbols used in an expression, doesn't provide inter-node relationships and may lead to mismatch.
To make abstract mathematical expressions feasible for computation and generation, we take advantage of \textit{symbolic expression trees} \cite{petersen2019deep}, which is a binary tree with nodes representing either operators or operands. 

\noindent\textbf{Symbol Library}. 
First, we need to establish a symbolic library consisting of candidate operators and operands for expression generation. 
During the generation process, the generator iteratively selects operators or operands---called tokens---from the library to construct a symbolic expression tree. 
We use $\mathcal{L}$ to denote the search token library, including operator sets $\{+, -, \times, \div, \log, \exp, pow\}$, operand set $\{x_1, x_2, \dots, x_k\}$ and constant set $\{0.2,0.5,2.0,5.0\}$.
In practice, neither the introduction of more complex operators such as $\{\sin,\cos\}$ nor the refinement of constants can achieve better performance. The comparative results can be seen in Appendix C.

For each child node in the expression tree, we can obtain the sub-tree that roots at this node.
Later, the expression corresponding to this sub-tree forms the argument of the parent node.
Some operators, such as addition, subtraction, multiplication and division, receive two input arguments and thus have two children in the expression tree.
Other operators, such as $\log$, $\exp$, and $pow$, receive a single input argument and thus have only one child in the expression tree.
The emergence of an operand node or constant node indicates the termination of the current branch since an operand contains no argument and must be the leaf node. 

Finally, we use pre-order traversal-generated sequences to represent a symbolic expression tree. 
The pre-order traversal sequences not only retain a one-to-one correspondence to an expression but also contain enough information on parents and siblings, which is crucial for generation models relying on the context.

\subsection{Generation of Mathematical Expressions}
In this part, we will introduce the generation process of the mathematical expressions. 
Specifically, we first introduce the input features of the expressions, i.e., the design of operand tokens we use to perform symbolic regression.
Second, we formulate the expression generation process as a sequence generation process and parameterize the generator as a sequence model.
Third, we introduce the RL-based training algorithm for the generator.

\subsubsection{Input Features}
We utilize the features designed in \cite{he2014learning} as input operands: (a) current node information including lower bound, estimated objective, depth, and the relationship to the last selected node; (b) branching features including pseudocost, gap and current bound; (c) B\&B tree state including global bounds, integrality gap, whether the gap is finite, and number of found solutions. 
Most solvers record all the features we use. 
Thus, it is feasible for training and deployment on different solvers using this set of designed features.
For more details on the features, please see Table \ref{tab:feat} in Appendix \ref{features}.

\subsubsection{Sequential Model}
We can employ a sequential model to generate desired expressions using the traversal sequence representation. 
When sampling a node of the symbolic expression tree, the parent and the sibling nodes serve as input to a recurrent neural network (RNN) to generate a categorical distribution associated with token library $\mathcal{L}$. 
We then sample according to this distribution.
Thus, the likelihood of the entire expression sequence $\tau$ is as follows:
\begin{align*}
    p_{\mathbf{\theta}}(\tau)=\prod^{|\tau|}_{i=1}p_{\mathbf{\theta}}(\tau_i\ |\ \tau_{1:(i-1)}),
\end{align*} 
where $\tau_i$ is the $i^{th}$ token in traversal sequence and $\mathbf{\theta}$ are the parameters of RNN. 

\subsubsection{Reinforcement Learning Formulation}
To tackle the complex and discrete symbolic space, a straightforward approach to optimize the parameters $\theta$ is exploring the search space with reinforcement learning (RL). The detailed methodology of RL can be found in Appendix B.3.
To formulate the RL problem, we view the expression generator as the agent and the generation strategy as a state-to-action policy $p_{\mathbf{\theta}}(\tau)$. 
Then, we specify the state, action and reward in the RL formulation as follows.

\noindent \textbf{State}. 
The states represent the current context or information available to the RNN at each time step. 
For sequence generation, a state is the parent and sibling nodes $(parent, sibling)$ of the current token. 
If the current token does not have any parent node (the root node) or the siblings have not been generated yet, we would put a blank state to the corresponding place as a placeholder.
The state would typically include the information on previously generated expression sequences.

\noindent \textbf{Action}. 
The actions represent the decisions made by the RNN at each time step, which is the next token in the sequence, i.e. $\tau_i\in\mathcal{L}$. 
The action space is the token library $\mathcal{L}$.
The logits predicted by RNN for each token can be considered as a probability distribution over possible actions.

\noindent \textbf{Reward}. 
The reward signal provides feedback to the RNN about the quality of the generated sequences. 
Since we only evaluate complete expressions, the reward remains zero until the generation for the expression is finished. 
The reward could be based on various criteria, such as how well the generated sequence matches a target sequence, how coherent the sequence is, or any other task-specific objective. 
Here we use the \textit{Fitness Measure} described in the next part.

At each time step $t$, the agent observes the current state $s_t$ and selects an action $a_t\in \mathcal{A}$ according to the policy.
Then, the agent takes this action and gains a reward $r_t\in \mathbb{R}$ from the environment. 
The objective of reinforcement learning is to find an optimal policy that maximizes the expected reward over time.

\subsection{Fitness Measure}
A natural idea to evaluate the generated expression is to use the end-to-end solving time of solvers with the expressions.
However, the solving time is not a stable choice for training, because the time fluctuates highly relating to quality and current usage of CPU.
A better alternative is the behavioral cloning accuracy with respect to the expert selecting policy, which takes the form of:
\begin{align*}
r(\tau)=\mathbb{E}_{((n^{(1)},n^{(2)}), d)\sim D}\left[
    \mathbb{I}\left(
    \tau(n^{(1)},n^{(2)})=d
    \right) \right]
\end{align*}
where $D=\{((n^{(1)}_t,n^{(2)}_t), d_t)\}_{t=1}^T$ is the sampled data in the node selection process including node features $n^{(1)}_t$ and $n^{(2)}_t$, expert decision $d_t$ of  current active node pairs.
The indication function $\mathbb{I}(\cdot)$ returns 1 if and only if $\cdot$ is true, which represents the node with a higher expression score coincides with the expert decision. 
Finally, the expectation measures how exactly the expression follows the expert decisions along the solving trajectory.

\subsection{Training}

\noindent\textbf{Policy Gradient Loss}. The training objective is to optimize a parametric policy $p_{\mathbf{\theta}}(\tau)$ and maximize the expected fitness,
\begin{align*}
    \mathop{\max}\limits_{\mathbf{\theta}} \mathbb{E}_{\tau\sim p_{\mathbf{\theta}}(\tau)}\left[r(\tau)\right]
\end{align*}
The common policy gradient objective defined above aims to optimize the average performance, i.e. expressions with poor effect are also considered to improve.
Instead, here we employ the risk-seeking objective from \cite{petersen2019deep} that focuses on the top-$\epsilon$ performance sequences to generate the best-fit expression, where the objective takes the form as follows:
\begin{align*}
    J_{risk}(\mathbf{\theta};\epsilon)=\mathbb{E}_{\tau\sim p_{\mathbf{\theta}}(\tau)}\left[r(\tau)\ |\ r(\tau)\geq r_\epsilon(\mathbf{\theta})\right],
\end{align*}
where $r_\epsilon(\mathbf{\theta})$ is the $(1-\epsilon)$ quantile of all the rewards in the current batch.
The \textit{risk} actually derives from the defects in the control environment model produced by only concentrating on performance in a certain aspect. 
For the generation, we seek risks to get the best expression while wiping out all the worst.

\noindent\textbf{Proximal Policy Optimization}. The proximal policy optimization (PPO) proposed by \cite{schulman2017proximal} is a sample-efficient reinforcement learning algorithm. Denote $q(\mathbf{\theta})=\frac{p_{\mathbf{\theta}}(\tau)}{p_{\mathbf{\theta}_{old}}(\tau)}$ as the probability ratio of the current and the old policies. To restrict the range of updating, the clip function is given by:
\begin{align*}
    q^{clip}_\eta(\mathbf{\theta})=\left\{
    \begin{aligned}
        &1+\eta,\quad\hat{A}>0\ and\ q(\mathbf{\theta})\geq 1+\eta \\
        &1-\eta,\quad\hat{A}<0\ and\ q(\mathbf{\theta})\leq 1-\eta \\
        &q(\mathbf{\theta}),\quad otherwise
    \end{aligned}
    \right .
\end{align*}
where $\hat{A}=r(\tau)-V_\alpha(\tau)$ is an estimator of the advantage, $V_\alpha(\tau)$ is the baseline state-value function. Eventually, PPO optimizes:
\begin{align*}
    J_{risk}(\mathbf{\theta};\epsilon, \eta) = \mathbb{E}_{\tau\sim p_{\mathbf{\theta}}(\tau)}\left[q^{clip}_\eta(\mathbf{\theta})\hat{A}\ |\ r(\tau)\geq r_\epsilon(\mathbf{\theta})\right]
\end{align*}
Furthermore, \textit{the hierarchical entropy regularizer} and \textit{soft length regularizer} \cite{landajuela2021improving} are introduced to reach a more efficient search in expression space, as hierarchical entropy regularizer mitigates the tendency of selecting paths with the same initial branches and soft length regularizer revises the heavy skew toward longer expressions.

\begin{table*}[htb]
\centering
\label{tab:t1}
\caption{
Comparison of average solving performance between our approach and baseline methods, under a $1,000$-second time limit.
`Expert' policy utilizes the solution to select nodes.
`Times' represents the time cost for solving an instance.
`PDI' denotes the integral of the primal and dual bound curves along time.
`Nodes' reflects the size of the final B\&B search tree and is CPU-independent.
We mark \textbf{the best} values in bold and underline \underline{the second-best} values while `$\downarrow$' indicates that lower is better intuitively.
Notice that GNN and Ranknet are GPU-based approaches.}
\setlength{\tabcolsep}{1.7mm}{
\fontsize{8.5pt}{11pt}\selectfont{
\begin{tabular}{@{}cccccccccc@{}}

\toprule
FCMCNF & \multicolumn{3}{c}{Easy  } & \multicolumn{3}{c}{Medium  } & \multicolumn{3}{c}{Hard  } \\ 
Model & Times$(s)$ $\downarrow$& PDI$\downarrow$ & Nodes$\downarrow$ & Times$(s)$ $\downarrow$ & PDI$\downarrow$ & Nodes $\downarrow$& Times$(s)$ $\downarrow$ & PDI$\downarrow$ & Nodes$\downarrow$ \\ 
\midrule
Expert & $3.43_{\pm 1.45}$ & 88.03 & $32.98_{\pm 3.12}$ & $18.92_{\pm 1.51}$ & 512.30 & $79.83_{\pm 3.53}$ & $77.91_{\pm 1.99}$ & 2243.31 & $240.66_{\pm 3.07}$ \\
\cmidrule(lr){1-4}
\cmidrule(lr){5-7}
\cmidrule(lr){8-10}
DFS & $4.07_{\pm 1.48}$ & \textbf{94.41} & $82.64_{\pm 3.33}$ & $25.17_{\pm 1.57}$ & \textbf{559.28} & $283.35_{\pm 2.47}$ & $113.28_{\pm 1.92}$ & \textbf{2643.18} & $575.85_{\pm 2.32}$ \\
Estimate & $4.11_{\pm 1.51}$  & \underline{94.87} & $85.97_{\pm 3.45}$ & $27.04_{\pm 1.62}$ & \underline{577.98} & $281.54_{\pm 2.24}$ & $123.00_{\pm 2.07}$ & \underline{3005.19} & $486.56_{\pm 2.47}$ \\
\cmidrule(lr){1-4}
\cmidrule(lr){5-7}
\cmidrule(lr){8-10}
SVM        & $3.98_{\pm 1.46}$ & 96.68 & $62.44_{\pm 3.18}$ & $\underline{23.38}_{\pm 1.65}$ & 645.82 & $172.40_{\pm 3.57}$ & $\underline{110.45}_{\pm 2.27}$ & 3756.57 & $\textbf{463.41}_{\pm 3.39}$ \\
Ranknet    & $3.98_{\pm 1.48}$  & 99.79 & $\underline{59.66}_{\pm 3.22}$ & $24.92_{\pm 1.87}$ & 659.17 & $175.22_{\pm 4.51}$ & $121.15_{\pm 2.38}$ & 3935.91 & $533.67_{\pm 3.45}$ \\
GNN        &  $\underline{3.96}_{\pm 1.46}$  & 97.50  & $67.43_{\pm 2.92}$ & $23.76_{\pm 1.78}$ & 679.64 & $\underline{168.16}_{\pm 3.72}$ & $\textbf{110.14}_{\pm 2.20}$ & 3149.55 & $\underline{484.91}_{\pm 3.64}$ \\
Dso4NS  & $\textbf{3.81}_{\pm 1.47}$ & 96.59 & $\textbf{54.16}_{\pm 3.05}$ & $\textbf{22.12}_{\pm 1.63} $& 669.50 & $\textbf{146.44}_{\pm 3.53}$ & $113.60_{\pm 2.26}$ & 4325.54 & $492.58_{\pm 3.32}$ \\
\bottomrule
\toprule
facilities & \multicolumn{3}{c}{Easy  } & \multicolumn{3}{c}{Medium  } & \multicolumn{3}{c}{Hard  } \\ 
Model & Times$(s)$ $\downarrow$& PDI$\downarrow$ & Nodes$\downarrow$ & Times$(s)$ $\downarrow$ & PDI$\downarrow$ & Nodes $\downarrow$& Times$(s)$ $\downarrow$ & PDI$\downarrow$ & Nodes$\downarrow$ \\ 
\midrule
Expert & $44.05_{\pm 2.73}$ & 449.84 & $134.13_{\pm 5.04}$ & $45.59_{\pm 2.04}$ & 540.47 & $105.82_{\pm 3.48}$ & $136.66_{\pm 1.70}$ & 1294.56 & $155.56_{\pm 2.88}$ \\
\cmidrule(lr){1-4}
\cmidrule(lr){5-7}
\cmidrule(lr){8-10}
DFS & $79.33_{\pm 2.79}$  & 642.71 & $272.15_{\pm 4.26}$ & $77.67_{\pm 1.79}$ & \textbf{599.95} & $295.87_{\pm 2.52}$ & $265.19_{\pm 1.68}$ & \textbf{1557.29} & $360.24_{\pm 2.06}$  \\
Estimate & $91.53_{\pm 2.47}$ & \textbf{598.58} & $332.08_{\pm 3.83}$ & $96.04_{\pm 2.01}$ & \underline{678.55} & $326.71_{\pm 2.72}$ & $288.87_{\pm 2.10}$ & 1950.79 & $347.78_{\pm 2.39}$ \\
\cmidrule(lr){1-4}
\cmidrule(lr){5-7}
\cmidrule(lr){8-10}
SVM        & $\textbf{51.23}_{\pm 2.70}$ & 637.00 & $\underline{167.90}_{\pm 5.03}$ & $63.73_{\pm 2.03}$ & 781.05 &$ 201.72_{\pm 3.52}$ & $199.78_{\pm 1.60}$ & 1812.63 & $299.29_{\pm 2.37}$ \\
Ranknet    & 4$8.89_{\pm 2.72}$ &  \underline{634.29} & $\textbf{156.92}_{\pm 5.19}$ & $60.48_{\pm 1.96}$ & 795.04 & $210.76_{\pm 3.17}$ & $192.71_{\pm 1.50}$ & 1808.75 & $\underline{274.54}_{\pm 2.56}$ \\
GNN        &  $52.88_{\pm 2.86}$ & 650.32 & $175.64_{\pm 4.95}$ & $\underline{60.41}_{\pm 2.23}$ & 787.34 & $\underline{192.27}_{\pm 3.78} $& $\underline{186.05}_{\pm 1.54}$ & \underline{1697.79} & $293.75_{\pm 2.46}$  \\
Dso4NS  & $\underline{51.64}_{\pm 2.76}$ & 666.92 & $179.42_{\pm 4.69}$ & $\textbf{59.23}_{\pm 2.15}$ & 815.25 & $\textbf{184.88}_{\pm 3.59}$ & $\textbf{177.28}_{\pm 1.58}$ & 1718.55 & $\textbf{260.04}_{\pm 2.76}$  \\
\bottomrule
\toprule
setcover & \multicolumn{3}{c}{Easy  } & \multicolumn{3}{c}{Medium  } & \multicolumn{3}{c}{Hard  } \\ 
Model & Times$(s)$ $\downarrow$& PDI$\downarrow$ & Nodes$\downarrow$ & Times$(s)$ $\downarrow$ & PDI$\downarrow$ & Nodes $\downarrow$& Times$(s)$ $\downarrow$ & PDI$\downarrow$ & Nodes$\downarrow$ \\ 
\midrule
Expert & $11.81_{\pm 1.75}$ & 651.01 &$ 70.97_{\pm 5.57}$ &$12.36_{\pm 1.54}$ &642.33  &$110.72_{\pm 4.37}$ &$96.69_{\pm 2.22}$ &2311.40 &$2825.55_{\pm 3.62}$ \\
\cmidrule(lr){1-4}
\cmidrule(lr){5-7}
\cmidrule(lr){8-10}
DFS  & $14.37_{\pm 1.78}$ & \textbf{685.05} & $148.98_{\pm 4.87}$ & $14.83_{\pm 1.58}$ & \textbf{659.47} &$187.96_{\pm 3.74}$ &$\textbf{101.76}_{\pm 2.13}$ &\textbf{2169.73} &$\underline{3528.19}_{\pm 2.98}$  \\
Estimate & $16.81_{\pm 2.02}$ & \underline{721.67} & $189.42_{\pm 5.52}$ & $16.75_{\pm 1.75}$ & \underline{672.05} &$227.18_{\pm 3.97}$  &$165.17_{\pm 2.27}$ &3504.92 &$4474.54_{\pm 3.04}$ \\
\cmidrule(lr){1-4}
\cmidrule(lr){5-7}
\cmidrule(lr){8-10}
SVM     & $13.26_{\pm 1.82}$ & 733.96 & $98.33_{\pm 5.52}$ & $\textbf{13.49}_{\pm 1.58}$ &704.04 &$\underline{131.92}_{\pm 4.40}$ &$105.29_{\pm 2.25}$ &2751.79 &$3673.52_{\pm 3.37}$ \\
Ranknet  & $13.19_{\pm 1.79}$ & 731.61 & $99.96_{\pm 5.25}$ & $\underline{13.53}_{\pm 1.57}$ &706.20  &$138.24_{\pm 4.29}$ &$108.71_{\pm 2.34}$ &2832.13 &$3767.54_{\pm 3.79}$ \\
GNN      & $\underline{13.16}_{\pm 1.80}$ & 732.36 & $\underline{95.24}_{\pm 5.48}$ & $13.67_{\pm 1.57}$ &707.23 & $142.06_{\pm 4.34}$ &$\underline{103.46}_{\pm 2.12}$ &2887.34 &$3567.86_{\pm 2.89}$ \\
Dso4NS  & $\textbf{13.11}_{\pm 1.77}$ & 724.94 & $\textbf{87.74}_{\pm 5.14}$ & $13.68_{\pm 1.74}$ & 702.68 & $\textbf{124.38}_{\pm 5.30}$ &$104.47_{\pm 2.24}$ &\underline{2749.28} &$\textbf{3438.73}_{\pm 3.52}$ \\
\bottomrule
\end{tabular}}}
\label{tab:main-results}
\vspace{-2mm}
\end{table*}

\begin{algorithm}
    \caption{Deep Symbolic Optimization for Node Selection}
    \label{alg1}
    \begin{algorithmic}
        \State // Behaviour generation
        \State  \textbf{Input: } the library of tokens $\mathcal{L}$, the sequential model with parameters $\theta$, the $M$ instances of MILP problems with solutions
        \State  Initial expert behavior buffer: $D\leftarrow\emptyset$
        \For{instance $= 1 \to M$}
            \While {solving is not complete}
                \State  Record state feature $n^{(1)}_t,n^{(2)}_t$ of two nodes being compared and expert decision $d_t$ at step $t$
            \EndWhile
            \State  Collect expert behaviors: $D\leftarrow D \cup\{(n^{(1)}_t,n^{(2)}_t, d_t)\}^T_{t=1}$
        \EndFor
        \State // Train the sequential model
        \For{iteration $=1 \to N$}
            \State Sample $K$ symbolic functions from token distribution $\tau_i\sim p_{\theta|\tau_{i-1}}$, where $\tau_i \in \mathcal{L}$
            \For{$j = 1 \to K$}
                \State $r(\tau^j)=\mathbb{E}_{(n^{(1)},n^{(2)}, d)\sim D}\left[\mathbb{I}(\tau^j(n^{(1)},n^{(2)})=d)\right]$
            \EndFor
            \State Train model: update $\theta$ via PPO by optimizing $J_{risk}(\mathbf{\theta};\epsilon, \eta)$
        \EndFor
        \State Symbolic function $\tau$ with highest $r(\cdot)$ on validation set
    \end{algorithmic}
    \vspace{-1mm}
\end{algorithm}

\subsection{Deployment}
After finishing the training process, we select the symbolic policy $\tau_{best}$ with the highest accuracy on the validation set. 
The symbolic policy compares all the active nodes two-by-two and selects the best one to explore next. 
Intuitively, this policy can be regarded as a data-driven version of the human-designed \textit{NODECOMP} integrated into modern CO solvers. 
Thus, we directly leverage the learned policy to replace the heuristics.
The pseudo-code of our algorithm is summarized in Algorithm \ref{alg1}. Our code is available at \url{https://github.com/sayal-k/dso4ns}.

\section{Results}

\subsection{Setup}
\subsubsection{Benchmarks}
We evaluate the performance of our approach on three NP-hard instance families.
The first benchmark is composed of Fixed Charge Multicommodity NetworkFlow (FCMCNF)
\cite{hewitt2010combining} instances. Following the \textit{FCMCNF} generation setting in \cite{labassi2022learning}, we generate train and test (Easy) instances with $n=15$ nodes and $m= 1.5·n$ commodities, yielding 354 constraints and 1401 variables on average. We further generate transfer instances with 20 (Medium), and 25 (Hard) nodes, separately yielding 640, 1016 constraints and 3378, 6718 variables on average.
The second benchmark is the capacitated facility location (facilities) proposed in \cite{cornuejols1991comparison}, which are challenging and representative of problems emerging in practice. We train and test (Easy) on instances with 100 customers, and generate transfer instances with 200 (Medium) and 400 (Hard) customers,  correspondingly yielding 10201, 20301, and 40501 constraints and 10100, 20100, and 40100 variables on average. 
The last benchmark is the set covering \cite{balas1980set} instances based on the family of cutting planes from conditional bounds. We generate train and test (Easy) with 500 constraints, and transfer sets with 1000 (Medium), 2000 (Hard) constraints, while variables are all set to 1000.

For each benchmark, we generate 1000 training instances, 100 validation instances, 30 test instances, 30 medium transfer instances and 30 hard transfer instances from open-source codes. 
Running behavior collection on training, validation and test instances yielded 63751 training, 3341 validation and 1973 test samples for FCMCNF, 85712 training, 8253 validation and 6380 test samples for facilities, 77958 training, 9298 validation and 4362 test samples for setcover.

\subsubsection{Baseline}
We compare against baselines including expert policy, heuristic selection rules and machine learning approaches. 
The expert policy reads the solution ahead and takes the form of \textit{optimal plunger}, which means it will plunge into the node whose constraints domain contains the value in the solution. 
Node selection heuristics comprise DFS and ESTIMATE implemented in SCIP. The DFS policy always selects the deepest among the promising nodes and ESTIMATE utilizes the estimated value of the best feasible solution in subtree of the node.
While in SCIP, the \textit{NODECOMP} heuristic is implemented in conjunction with the diving \textit{NODESELECT} heuristic.
We follow the setting of SCIP solver in \cite{labassi2022learning} to disentangle the effect brought by the diving \textit{NODESELECT} rule.
Specifically, we make a comparison to default SCIP (SCIP) and heuristics with a plain \textit{NODESELECT} rule that always selects the node with the highest score (ESTIMATE).

We use three machine learning baselines as follows. 
\cite{he2014learning} leverages the support vector machine (SVM) for node selection. 
The multilayer perceptron approach Ranknet of \cite{song2018learning} and \cite{yilmaz2021study} rank the two given nodes and select a superior one. 
Here we implement Ranknet using 2 layers with 50 hidden neurons and utilize the fixed-dimensional features of \cite{he2014learning} for SVM and Ranknet. 
The graph neural network (GNN) approach proposed by \cite{labassi2022learning} takes the graph representation of nodes in the search tree as input to a GNN scoring function. 
Note that we run the Ranknet and GNN on a GPU machine, and run the other methods on a single CPU. 
All the ML methods use the aforementioned plain \textit{NODESELECT} rule.

\begin{table*}[t]
    \centering
    \caption{
    Training Efficiency comparison. `-1' means model is trained under behaviors of one instance. `Dso4NS-1-symm' refers to the attempt to learn a self-consistent scoring function with features of one node as input, while `Dso4NS-1' leverages features of two nodes to directly make decisions. Here we mark \textbf{the best} values among `-1' models in bold.
    }
    \scalebox{1}{\setlength{\tabcolsep}{1.7mm}{
\fontsize{8.5pt}{11pt}\selectfont{
\begin{tabular}{@{}cccccccccc@{}}
\toprule
facilities & \multicolumn{3}{c}{Easy  } & \multicolumn{3}{c}{Medium  } & \multicolumn{3}{c}{Hard  } \\ 
Model & Times$(s)$ $\downarrow$& PDI$\downarrow$ & Nodes$\downarrow$ & Times$(s)$ $\downarrow$ & PDI$\downarrow$ & Nodes $\downarrow$& Times$(s)$ $\downarrow$ & PDI$\downarrow$ & Nodes$\downarrow$ \\ 
\midrule
SVM-1        & 46.88 & 548.54 & 162.67 & 79.97 & 614.40 & 550.74 & 189.24 & 1886.27 & 278.98 \\
Ranknet-1    & 47.45 & 536.33 & 164.58 & 77.17  &  600.02 & 534.70 & 187.09 & 1861.42 & 277.93 \\
GNN-1        & 47.23 & 553.85 & 164.89 &  79.80   & 615.32 & 551.88  & 189.77 & 1892.85 & 281.22 \\
\cmidrule(lr){1-4}
\cmidrule(lr){5-7}
\cmidrule(lr){8-10}
Dso4NS-1       & 45.07 & 521.50 & \textbf{140.90}  & 77.21 & 598.04 & 526.18 & 189.72 & 1871.67 & 280.92  \\
Dso4NS-1-symm   & \textbf{43.73} & \textbf{515.99} & 147.55 & \textbf{74.34} & \textbf{592.80} & \textbf{480.64} & \textbf{186.48} & \textbf{1835.10} & \textbf{277.02}  \\
\midrule
Dso4NS-1000       & 45.38 & 523.84 & 170.19 & 72.98 & 588.08 & 467.97 & 184.67 & 1874.08 & 273.08  \\
Dso4NS-1000-symm   & 45.13 & 521.35 & 174.49 & 74.71 & 590.65 & 489.50 & 188.90 & 1884.55 & 281.78 \\
\bottomrule
\end{tabular}}}}
\label{tab:train}
\vspace{-1mm}
\end{table*}

\subsubsection{Training Settings}
We conducted all the experiments on a single machine with NVidia GeForce GTX 3090 GPUs and Intel(R) Xeon(R) E5-2667 v4 CPUs @3.20GHz. The SVM is implemented by the scikit-learn \cite{pedregosa2011scikit} library; the Ranknet and GNN are based on PyTorch \cite{paszke2019pytorch} and optimized using Adam\cite{kingma2014adam} with BCELoss and training batch size of 16. Our model is composed of 2 LSTM layers with a hidden size of 128, trained under 0.2 risk factor, hierarchical entropy coefficient of 0.9 and PPO learning rate of 5e-5, optimized using Adam with a batch size of 500 under 300 iterations. 
In real-world scenarios, solvers encounter problems with similar structures and are accustomed to addressing analogous problems; hence, we conducted individualized training for each benchmark. Furthermore, we limit the maximum number of tokens to 10 and select the best-performed expression according to the accuracy on the validation set to reduce overfitting.
Our code for symbolic regression is based on the open-source DSO framework \url{https://github.com/dso-org/deep-symbolic-optimization}.

\subsubsection{Evaluation}
The state-of-the-art open-source solver SCIP acts as the backend solver through all the evaluations, where we set the solving time limit to 1000s and deactivate the pre-solve and heuristic features. 
We evaluate each ML model by deploying the model-based scoring function into the \textit{NODECOMP} module. 
We report results using the average running time in seconds (Time), the hardware-independent B\&B tree size (Nodes), and the integral of the primal and dual bound curves along time (PD Integral). 
The \textit{Time} metric directly reflects the solving efficiency for which bad policies cannot obtain solutions and reach the time limit. 
The \textit{PD Integral} metric counts on not only time but also the shrinkage speed of bounds, indicating whether the node chosen by the policy is more effective. 
While time metrics are heavily relative to CPU load thus leading to mismatch, the \textit{Nodes} metric reflects the exploring performance for better policies heading towards the solution, thus exploring fewer nodes in the search tree.
In light of this, we keep a low CPU load during evaluation to make a fair comparison on the Time metric.
To decide the performance of an approach, we focus on the solving time of the baselines, where a shorter solving time indicates superior performance.

\begin{table}[h]
    \centering
    \caption{
    The average decision-making time of ML-based policies in milliseconds, including costs of feature extraction and inference. 
    }
    \scalebox{0.8}{
        \begin{tabular}{ccccc}
        \toprule
            FCMCNF & SVM & Ranknet & Dso4NS & GNN\\
            \midrule
            Easy & 0.34 & 15.16 &  0.07 & 2.49\\
            Medium & 0.45 & 25.82 & 0.13 & 2.54\\
            Hard & 0.44 & 15.01 & 0.10 & 2.64\\
        \bottomrule
        \toprule
            facilities & SVM & Ranknet & Dso4NS & GNN\\
            \midrule
            Easy & 0.61 & 15.27 & 0.14 & 2.98\\
            Medium & 0.60 & 25.52 & 0.13 & 3.38\\
            Hard & 0.46 & 37.85 & 0.14 & 2.91\\
        \bottomrule
        \toprule
            setcover & SVM & Ranknet & Dso4NS & GNN\\
            \midrule
            Easy & 0.46 & 7.68  &  0.10 & 2.88 \\
            Medium & 0.47 & 13.25 & 0.09 & 2.73\\
            Hard & 0.44 & 20.43 & 0.10 & 2.74\\
        \bottomrule
        \end{tabular}
    }    
\label{tab:inf}
\vspace{-2mm}
\end{table}

\subsection{Comparative Experiments}
\subsubsection{Solving Efficiency} We compare the solving performance of Dso4NS learned policies to baselines, and report results in Table \ref{tab:main-results}. Results show that Dso4NS policies achieve expected performance and outstand other baselines in most cases, especially the \textit{Nodes} metrics. Despite the specific optimization to DFS on setcover (a most common benchmark) in SCIP, the performance degradation of Dso4NS on FCMCNF hard transfer may be due to the limitation of expression sequence length, since fewer input features constraints decision-making ability in complex situations and the learning capability.

\subsubsection{Inference Efficiency} We compare the inference efficiency of different approaches with the time cost for single feature extraction and inference, named the decision-making time.  Results in Table \ref{tab:inf} demonstrate that compared to SVM with 40 dimensions input and GPU accelerated GNN, Dso4NS is at an extremely low time cost, which reduces the total solving time when searching more nodes than other methods.

\subsubsection{Node Selection Accuracy} We evaluate the accuracy under the expert selection behaviors, reported in Table \ref{tab:acc}. Concerning the metrics above, Ranknet with the highest accuracy doesn't exhibit the best performance and GNN with the lowest accuracy reaches the second-best sometimes. The accuracy does not show a strong correlation with the solving efficiency because of cumulative wrong selection bias and termination conditions set by solvers. But Dso4NS's high accuracy still shows its learning capability compared to other ML approaches.
\begin{table}[h]
    \centering
    \caption{
    Selection accuracy with respect to expert behaviors.
    }
     \scalebox{0.8}{
        \begin{tabular}{cccccc}
        \toprule
            FCMCNF & SVM & Ranknet & GNN & Dso4NS\\
            \midrule
            Easy & 0.946 & 0.979 & 0.937 & 0.978 \\
            Medium & 0.969 & 0.989 & 0.961 & 0.991 \\
            Hard & 0.976 & 0.987 & 0.972 & 0.982 \\
        \bottomrule
        \toprule
            facilities & SVM & Ranknet & GNN & Dso4NS\\
            \midrule
            Easy & 0.985 & 0.990 & 0.974 & 0.990 \\
            Medium & 0.979 & 0.987 & 0.973 & 0.987 \\
            Hard & 0.972 & 0.988 & 0.960 & 0.986 \\
        \bottomrule
        \toprule
            setcover & SVM & Ranknet & GNN & Dso4NS\\
            \midrule
            Easy & 0.974 & 0.981 & 0.963 & 0.977 \\
            Medium & 0.979 & 0.985 & 0.970 & 0.982 \\
            Hard & 0.991 & 0.996 & 0.978 & 0.994 \\
        \bottomrule
        \end{tabular}
    }
\label{tab:acc}
\vspace{-2mm}
\end{table}

\subsubsection{The Learned symbolic expressions} We report the best symbolic policies in Table \ref{tab:itp} and the descriptions of features are listed in \ref{tab:feat}. Considering that the simple Ranknet with only 2 linear layers can also achieve good performance, the expressions are expected to be compact and effective. Results demonstrate the generated policies are brief and use only a few features respectively derived from the first 20 (node1) and the last 20 features (node2), which conforms to the outstanding inference efficiency aforementioned and the intuition of node selection. 

Furthermore, by observing that the expressions do not exhibit good self-consistency (the features input are not symmetric), we conduct the experiment again with only features of a node as input to generate a scoring function $g$ and make decisions by $\sigma(g(node1)-g(node2))$. Solving performance results are shown in \ref{tab:train} and discussions are in section 5. 

All the expressions exhibit great conciseness and interpretability, and we expect solver researchers to further optimize heuristic rules by mining the hidden patterns from them.

\begin{table}[h]
    \centering
    \renewcommand\arraystretch{1.2}
    \caption{
    Dso4NS learned symbolic expressions for each benchmark with different training set sizes. \textit{Symm} means it takes the same features of each node as input while \textit{non-symm} expressions may gain different results by switching the order of input. The details of input variables can be found in Appendix A.
    }
    \scalebox{0.9}{
    \begin{tabular}{ccc}
    \toprule
    \multirow{2}{*}{FCMCNF\_1}        & non-symm & $-x_{26}-x_{34}+x_{6}$ \\
    \cline{2-3}
                                      & symm     & $x_8*(-x_6 + x_8 + exp(x_2 + x_8))$\\ 
    \midrule
    \multirow{2}{*}{FCMCNF\_1000}     & non-symm & $x_{19} - x_{29} - x_{39}$ \\
    \cline{2-3}
                                      & symm     & $x_{16}*x_9/x_{19} - x_{19} - exp(x_8)$ \\ 
    \midrule
    \multirow{2}{*}{facilities\_1}    & non-symm & $x_{19} - x_{39} + 0.2$    \\
    \cline{2-3}
                                      & symm     & $x_{11} - (x_{19} + x_3 + x_7 + x_8)^2$ \\ 
    \midrule
    \multirow{2}{*}{facilities\_1000} & non-symm & $-x_{26} -x_{29} + x_6$   \\
    \cline{2-3}
                                      & symm     & $x_{16}*x_{9}/x_{19}-x_{19}-exp(x_{8})$ \\ 
    \midrule
    \multirow{2}{*}{setcover\_1}      & non-symm & $-x_{28}* x_{29} + x_{9}$    \\
    \cline{2-3}
                                      & symm     &$x_{20} - x_8*(x_{19}^2 + x_7 - 0.5)$ \\ 
    \midrule
    \multirow{2}{*}{setcover\_1000}   & non-symm & $-x_{16} - x_{26} + x_{6} + x_{9} -0.5$   \\
    \cline{2-3}
                                      & symm     &$0.5*x_6*(x_{10}*x_{11}^2 - x_8)$ \\
    \bottomrule
    \end{tabular}
    }
\label{tab:itp}
\vspace{-2mm}
\end{table}

\subsubsection{Training Efficiency.} We report the results in Table \ref{tab:train} according to the performance of models trained with one instance on the facilities benchmark. 
Facilities instance yielded 32 samples.
Dso4NS reveals strong learning ability on a small-scale training set for its good performance on facilities than other methods.
Specifically because of the enforcing strong regularization and non-linearity from the symbolic operators, Dso4NS is more practical for real-world tasks, where we need to extract patterns from precious and limited training samples. 
Moreover, the self-consistent expressions reveal stronger stability and higher learning efficiency than non-symmetric expressions, while non-symmetric ones hold higher performance caps that can be achieved on a larger training set. 

\subsection{Ablation Study}
We conduct the ablation study to identify the influence of introducing extra mathematical operators or constant choices in Table \ref{tab:ablation}. The results indicate that the inclusion of operators $\{\sin, \cos\}$ or constants placeholder in the extended token library does not yield any additional performance improvement, as the newly learned expressions remain unchanged compared to those obtained using the default library. Intuitively, the introduction of complex operators leads to unnecessary vibration, while constant refinement exacerbates the overfitting of the training set.

\begin{table}[h]
    \centering
    \caption{
    Learned expressions under ablation libraries. 
    }
    \scalebox{1}{
        \begin{tabular}{cc}
        \toprule
            FCMCNF & Expressions\\
            \midrule
            Default & $x_{19} - x_{29} - x_{39}$\\
            Extra operators & $x_{19} - x_{29} - x_{39}$\\
            Const placeholder & $x_{19} - x_{29} - x_{39}$\\
        \bottomrule
        \end{tabular}
    }    
\label{tab:ablation}
\vspace{-3mm}
\end{table}

\section{Limitations and Future Work}
In this paper, we leverage deep symbolic optimization to enhance the performance of node selection heuristics.
However, certain limitations exist, such as the requirement for expert design of features and the necessity for individual training on different data sets.
In future work, we are going to learn more complex heuristics in the solver, discover more expressive and general symbolic expressions and deploy the symbolic optimization framework to more modules in the solvers. 
The Dso4NS has the potential to bridge the gap between learning-based methods and a broader range of real-world applications.

\section{Conclusion}
We propose a novel deep symbolic optimization framework (Dso4NS) for the node selection module in a B\&B solver.
Dso4NS aims to combine the advantages of both mathematical expressions and deep learning models while addressing their drawbacks.
Dso4NS leverages a deep model to guide the search in the high-dimension expression space and generate high-performance mathematical expressions for heuristics construction.
Experimental results demonstrate that Dso4NS is able to achieve promising performance, high interpretability, remarkable training and inference efficiency.

\section{Acknowledgements}
The authors would like to thank all the anonymous reviewers for their insightful comments.
This work was supported by National Nature Science Foundations of China grants U19B2044.


\bibliographystyle{ACM-Reference-Format}
\bibliography{SymbNodeSelection}

\begin{table*}[h]
    \centering
    \caption{The 20 features used to represent a node in B\&B tree, proposed in \cite{he2014learning}}
    \scalebox{0.8}{
        \begin{tabular}{lll}
        \toprule
            Index & Name & Description\\
            \midrule
            1/21 & GAPINF & Whether the lower bound is zero or the upper bound is infinite\\
            2/22 & GAP & The current gap $(\text{upperbound} - \text{lowerbound})/|\text{lowerbound}|$\\
            3/23 & GLOBALUPPERBOUNDINF & Whether upper bound is infinite\\
            4/24 & GLOBALUPPERBOUND & The value of the upper bound divided by the root lower bound\\
            5/25 & PLUNGEDEPTH & The current plunging depth (successive times, a child was selected as next node)\\
            6/26 & RELATIVEDEPTH & The depth of the node relative to the MAXDEPTH\\
            7/27 & LOWERBOUND & The lower bound of the node relative to the root lower bound\\
            8/28 & ESTIMATE & The estimated value of the best feasible solution in the sub-tree of the node\\
            9/29 & RELATIVEBOUND & The node lower bound gap relative to the current gap \\
            10-12/30-32 & NODE\_TYPE & Whether the node is sibling, child or leaf\\
            13/33 & BRANCHVAR\_BOUNDLPDIFF & \makecell[l]{The difference between the new value of the bound in the bound change data \\ and the current LP or pseudo solution value of variable}\\
            14/34 & BRANCHVAR\_ROOTLPDIFF & \makecell[l]{The difference between the solution of the variable in the last root node's relaxation \\and the current LP or pseudo solution value of variable}\\
            15/35 & BRANCHVAR\_PRIO\_DOWN & Whether the preferred branch direction of the variable is downwards\\
            16/36 & BRANCHVAR\_PRIO\_UP & Whether the preferred branch direction of the variable is upwards\\
            17/37 & BRANCHVAR\_PSEUDOCOST & The variable's pseudo cost value for the given step size "solvaldelta" in the variable's LP solution value\\
            18/38 & BRANCHVAR\_INF & The average number of inferences found after branching on the variable in given direction\\
            19/39 & NODE\_DEPTH & The depth of the node\\
            20/40 & MAXDEPTH & The depth of the node probing by solver\\
        \bottomrule
        \end{tabular}
    }    
\label{tab:feat}

\end{table*}

\newpage

\appendix

\section{Details of Input Features}

\label{features}
Here we list the corresponding description of each feature in Table \ref{tab:feat}.
The features are proposed in \cite{he2014learning}.
In our experiments, we denote the features of node 1 by $x[1-20]$ and node 2 by $x[21-40]$.

\section{Related works}
\subsection{Deep Symbolic Regression} 
To address the challenge of inefficiency in the large search space, researchers attempt to develop more effective search policies or prune the search space to reduce its complexity. 
Some recent works employ deep learning models to search for symbolic expressions, where the deep models incorporate historical information to guide effective searching.  
For example, \cite{petersen2019deep} develops a deep learning framework that uses the recurrent neural network (RNN) to generate symbolic sequences that represent expressions with a risk-seeking policy gradient. 
Then, they focus on and select the best-performing sequence samples. 
\cite{landajuela2021discovering} proposes deep symbolic policy (DSP) for control tasks.
DSP directly combines the generated symbolic policies and distilled pre-trained neural network-based policies to reduce the complexity of the search space.
Experiments show that DSP outperforms seven state-of-the-art deep reinforcement learning algorithms.
Recently, \citep{kuang2024rethinking} proposes the first deep symbolic branching framework for B\&B MILP solvers, in which the researchers leverage deep symbolic regression to learn a symbolic branching policy.
The learned simple symbolic expressions achieve a higher performance than the network policy.

\subsection{Machine Learning for Node Selection}
Existing works on ML for node selection aim to leverage machine learning models to replace the node selection heuristics in the solvers.
For example, they utilize support vector machine (SVM) \cite{he2014learning}, multi-layer perceptron (MLP), graph neural network (GNN), etc to learn a more efficient node selection policy. 
From the perspective of training methods, these works can be categorized into two classes: imitation learning (IL) and reinforcement learning (RL).

For IL methods, the deep model observes the expert's actions and tries to imitate the expert's decision behavior using behavior cloning techniques. 
During training, the optimal node selection trajectory, also known as the oracle, provides an optimal action $a^*$ for any possible state to be imitated by the policy. 
\cite{he2014learning} employs DAgger to support SVM on Node Selection, which iteratively updates the policy to align more closely with the expert's decisions encountered during previous runs of the policy. 
To exploit node features and global information effectively, \cite{labassi2022learning} develops a bipartite graph for each node, where they represent constraints and variables as vertices of the graph. 
A constraint vertex and a variable vertex are connected by an edge if the coefficient associated with the variable in the constraint is nonzero. 
Then the researchers parameterize the node selection policy as a GNN model and perform node selection by comparing the scores of each node computed by the 
 GNN. 
For RL methods, researchers design reward functions to explore and guide the learned policy in the direction of higher performance. 
\cite{zhang2023rl} proposes an RL approach to adapt to the dynamic nature of state and action spaces.
They try to combine and use the information of the early search tree and the cost associated with path backtracking.
In summary, the IL and RL methods for node selection inspired us that one can further improve the solver performance by incorporating distributional information for specific problems.
\vspace{-1mm}

\section{Reinforcement Learning}
Reinforcement learning is a computational approach that enables an \textbf{agent} to achieve its objectives through iterative interactions with the environment. 
In each interaction, the agent makes action decisions with its policy based on the current \textbf{state} of the environment, applies those actions, and receives feedback in terms of \textbf{rewards} and subsequent states. 
The goal of reinforcement learning is to maximize expected cumulative reward over multiple rounds of interaction. Unlike supervised learning's focus on models, reinforcement learning emphasizes that an agent not only perceives information about its environment but also actively influences it through decision-making rather than mere prediction signals.

\end{document}